\documentclass[aps,prx,reprint,superscriptaddress,longbibliography,nofootinbib,floatfix]{revtex4-2}

\usepackage{graphicx}
\usepackage{dcolumn}
\usepackage{bm}
\usepackage{amsmath}
\usepackage{amssymb}
\usepackage{physics}
\usepackage{hyperref}
\hypersetup{
    colorlinks=true,
    linkcolor=blue,
    citecolor=blue,
    urlcolor=blue
}

\begin{document}



\title{Noise-Driven Exploration and Transient Freezing Select Flat Minima in Stochastic Gradient Descent}

\author{Ning Yang}
\thanks{These authors contributed equally to this work.}
\affiliation{Peking University Chengdu Academy for Advanced Interdisciplinary Biotechnologies, Chengdu 610213, China}

\author{Yikuan Zhang}
\thanks{These authors contributed equally to this work.}
\affiliation{School of Physics, Peking University, Beijing 100871, China}

\author{Qi Ouyang}
\affiliation{Institute for Advanced Study in Physics, Zhejiang University, Hangzhou 310058, China}

\author{Chao Tang}
\affiliation{Peking University Chengdu Academy for Advanced Interdisciplinary Biotechnologies, Chengdu 610213, China}
\affiliation{Center for Quantitative Biology, Peking University, Beijing 100871, China}

\author{Yuhai Tu}
\email{ytu@flatironinstitute.org}
\affiliation{Center for Computational Biology \& Center for Computational Neuroscience, Flatiron Institute, New York, NY 10010, USA}

\begin{abstract}
    Stochastic gradient descent (SGD) is central to deep learning, yet the dynamical origin of its preference for flatter, more generalizable solutions remains unclear. Here, by analyzing SGD learning dynamics, we identify a nonequilibrium mechanism that governs solution selection during training. Numerical experiments reveal a transient exploratory phase in which SGD trajectories repeatedly escape sharp valleys and migrate toward flatter regions of the loss landscape before becoming confined to a final basin. Using a tractable physical model, we show that SGD noise reshapes the loss landscape into an effective potential that preferentially stabilizes flat solutions. We further uncover a transient freezing mechanism: as training progresses, the flattening landscape suppresses transitions between competing valleys. Stronger SGD noise delays this freezing transition, prolonging the exploratory phase and thereby increasing the probability of convergence to flatter minima. Together, these results provide a unified physical framework connecting learning dynamics, loss-landscape geometry, and generalization, and suggest guiding principles for the design of more effective optimization algorithms.

\end{abstract}

\maketitle

\section{Introduction}
Despite its success in a wide-range of applications, the underlying mechanisms by which deep learning selects a generalizable solution in the ultra-high dimensional weight space remain poorly understood \cite{lecunDeepLearning2015,levineMachineLearningMeets2024a}. Training a neural network can be viewed as optimization on a high-dimensional loss landscape $\mathcal{L}(\bm{\theta})$ driven by stochastic gradient descent (SGD) \cite{robbinsStochasticApproximationMethod1951,bottouLargeScaleMachineLearning2010}, where $\bm{\theta}$ denotes the trainable parameters \cite{liuLossLandscapesOptimization2022,sunGlobalLandscapeNeural2020}. A prevailing hypothesis is that the geometry of this landscape is closely linked to generalization: solutions lying in flat or wide minima tend to generalize better than those in sharp or narrow minima \cite{hochreiterFlatMinima1997,liVisualizingLossLandscape2018,jastrzebskiThreeFactorsInfluencing2018,wuHowSGDSelects2018,keskarLargeBatchTrainingDeep2017,chenSimpleConnectionLoss2024,fengActivityWeightDuality2023}. The central question is therefore not only why flat minima are favored statistically but also when and how the final solution is selected during the dynamical learning process. 

In the high-dimensional weight space, the loss landscape contains many competing solution valleys. Consistent with previous studies \cite{choromanskaLossSurfacesMultilayer2015,liVisualizingLossLandscape2018,fortLargeScaleStructure2019,garipovLossSurfacesMode2018,wuHowSGDSelects2018,ghorbaniInvestigationNeuralNet2019,chaudhariEntropySGDBiasingGradient2019}, we find that independent SGD runs initialized from the same point converge to multiple distinct low-loss endpoints (Fig.~S1 in the Supplemental Material). Although all these endpoints achieve zero training error, they form separated clusters and are divided by loss barriers along interpolation paths. Hessian spectra further reveal flat directions along the valley floor and steep transverse directions, supporting their interpretation as distinct solution valleys. Basin selection during training should therefore be understood as a dynamical process in which the optimizer navigates among competing valleys before committing to one of them.

The stochasticity driving this navigation has a distinctive structure. Due to mini-batch sampling, the noise covariance in SGD aligns with the local Hessian, making the noise both landscape-dependent and anisotropic \cite{zhuAnisotropicNoiseStochastic2019,xieDiffusionTheoryDeep2021,weiHowNoiseAffects2019,yangStochasticGradientDescent2023,xieOverlookedStructureStochastic2023,haochenShapeMattersUnderstanding2021,zhangSuperlinearRelationshipSGD2026}. This anisotropic stochasticity is believed to bias optimization trajectories toward flatter regions of the loss landscape \cite{xieDiffusionTheoryDeep2021,zhuAnisotropicNoiseStochastic2019,yangStochasticGradientDescent2023}. Empirical studies further indicate that increasing the overall noise strength of SGD---by raising the learning rate $\eta$ or reducing the batch size $B$---typically leads to convergence toward flatter minima with improved generalization \cite{keskarLargeBatchTrainingDeep2017,chaudhariEntropySGDBiasingGradient2019,jastrzebskiThreeFactorsInfluencing2018}. However, most existing analyses of this implicit regularization focus on late-stage or fixed-landscape regimes, where gradients are already small and trajectories have entered low-loss valleys. Such analyses do not, by themselves, explain how the final basin is selected during early training, when gradients remain large, barriers are still evolving, and the trajectory can remain mobile among competing valleys.

Multiple studies have demonstrated the critical importance of the early phase of training, showing that key properties of the final network are established long before convergence \cite{gur-ariGradientDescentHappens2018,achilleCriticalLearningPeriods2019,fengPhasesLearningDynamics2021a,kalraPhaseDiagramEarly2023,jastrzebskiBreakEvenPointOptimization2020}. For instance, the ``lottery ticket hypothesis'' suggests that trainable subnetworks emerge very early in the training process \cite{frankleLotteryTicketHypothesis2019}. This raises a crucial question: How does SGD navigate the complex, high-dimensional loss landscape during these formative early stages to set the trajectory toward generalizable solutions? Understanding this ``early transient dynamic'' is essential for a complete theory of deep learning optimization.



In this paper, we show that flat-minimum selection in SGD is governed by a nonequilibrium freezing process induced by Hessian-aligned anisotropic noise. Unlike a uniform thermal bath, SGD noise depends on the local landscape geometry: flatter valleys have weaker diffusion and therefore lower effective temperature. Equivalently, under a common effective-temperature representation, this anisotropic noise reshapes the loss into an effective potential in which flatter valleys have lower effective energy, which biases the inter-valley hopping towards the flatter valley. 
As training progresses, the loss landscape becomes increasingly flat, reducing the magnitude of SGD noise and progressively suppressing transitions between competing valleys. Eventually, inter-valley hopping freezes, trapping the dynamics within a single basin. Overall, the combination of self-cooling and freezing drives the SGD dynamics towards the flatter valleys. 


\section{Results}

\subsection{Transient hopping favors flatter valleys}

To study how SGD transitions between valleys during training directly, we designed a continuation training experiment. A network was first trained with SGD to generate a reference trajectory (Fig.~\ref{fig:phenomenology}(a); reference setting $B = 50$, $\eta = 0.05$). At selected checkpoint times $t_c$ along this trajectory, training was branched and resumed using deterministic full-batch gradient descent (GD). Because full-batch GD removes mini-batch stochasticity, each continuation rapidly descends into the local valley determined by its starting point, without further inter-valley exploration. The continuation endpoint therefore identifies the valley occupied by the SGD trajectory at time $t_c$ (Fig.~\ref{fig:phenomenology}(a); see Supplemental Material for details).

These results provide direct evidence that valley hopping during early training is biased toward flatter and more generalizable regions. Here, we quantify flatness $F$ as the inverse geometric mean of the top $N$ non-degenerate Hessian eigenvalues,
\begin{equation}
F \equiv \left( \prod_{i=1}^{N} \lambda_i(\mathbf{H}) \right)^{-\tfrac{1}{N}},
\label{eq}
\end{equation}
where we set $N=10$ for the 10-class MNIST task \cite{sagunEmpiricalAnalysisHessian2018a}. Continuations started from early times converge to solutions with relatively high test loss and low flatness. As $t_c$ increases, later continuations progressively land in valleys with lower test loss and higher flatness [Fig.~\ref{fig:phenomenology}(b), (c)]. In Fig.~\ref{fig:phenomenology}(c), flatness increases during deterministic continuation for all $t_c$ after the initial large-gradient regime. Note that the non-monotonic behavior of test loss---first decreasing and then rising---does not indicate worsening generalization; the test accuracy has already plateaued at that stage (Fig.~S3 in the Supplemental Material), and the late-time rise is an artifact of cross-entropy loss growing with increasingly extreme output logits. The flatness growth during continuation is partly attributable to the rescaling symmetry of ReLU networks~\cite{fengActivityWeightDuality2023}, which allows weight redistribution between layers without changing the network function. Because this rescaling affects all continuation trajectories similarly, it does not alter the fact that later starting times $t_c$ consistently yield flatter endpoints than earlier ones.

Importantly, continuations started from later times show faster flatness growth and therefore reach flatter final solutions. This systematic improvement indicates that the SGD trajectory does not remain trapped in its initial basin of attraction, but instead explores across the loss landscape to discover flatter and more generalizable regions.

\begin{figure*}[t]
    \centering
    \includegraphics[width=0.98\textwidth]{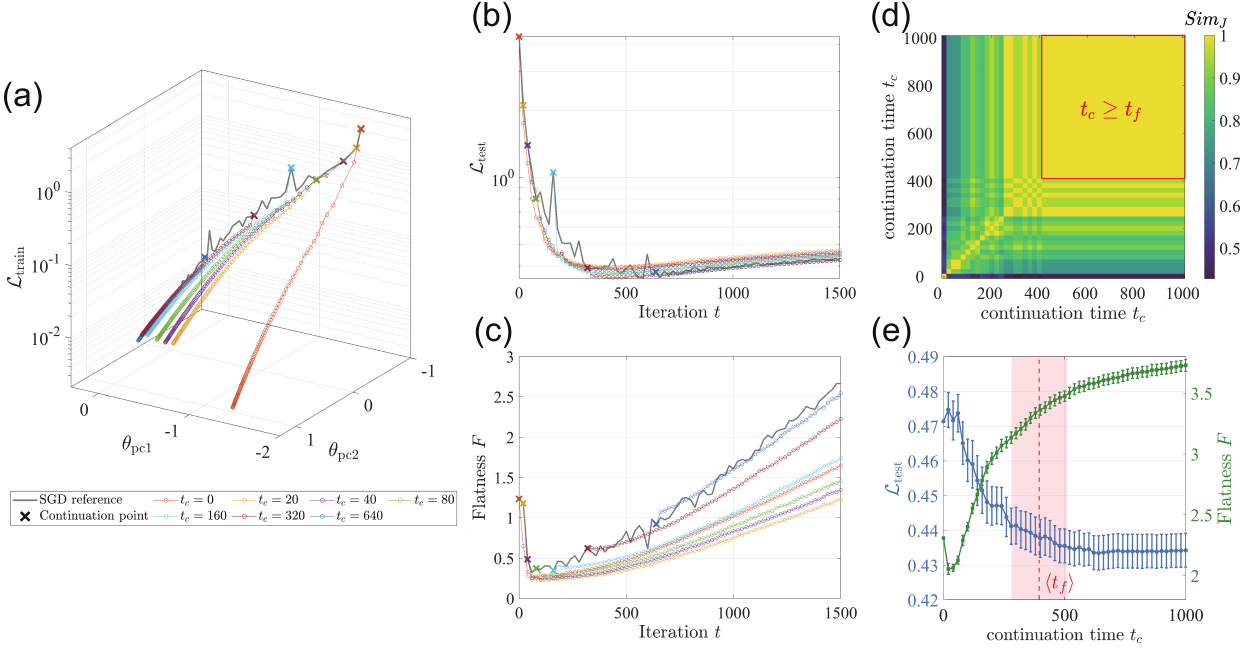}
    \caption{Continuation training reveals transient valley hopping and freezing during SGD. (a) A PCA projection of the reference SGD trajectory (gray; $B=50$, $\eta=0.05$) and deterministic full-batch GD continuations (colored curves) branched from different times $t_c$, with continuation points indicated by cross markers. The vertical coordinate indicates training loss. (b, c) Evolution of the test loss $\mathcal{L}_{\mathrm{test}}$ (b) and flatness $F$ (c) for the reference SGD trajectory and the GD-continued trajectories. (d) Jaccard similarity $Sim_J$ among solutions obtained from different continuation times $t_c$, with the red square marking the region where $t_c \geq t_{f}$. (e) Endpoint test loss and flatness (evaluated after deterministic continuation at $t=2000$) as functions of continuation time $t_c$, averaged over 20 independent repeats at $B=50$ and $\eta=0.05$; error bars denote the standard error of the mean (SEM). The vertical dashed line marks the mean freezing time $\langle t_f\rangle$, and the shaded band denotes $\langle t_f\rangle \pm \mathrm{SEM}$.}
    \label{fig:phenomenology}
    \end{figure*}

\subsection{The freezing transition}

In the latter stage of the exploration phase, we found that valley hopping stops at a well-defined freezing time. To identify when inter-valley mobility is lost, we compared the continuation endpoints obtained from different times. Let $\hat{\bm{\theta}}_i$ denote the endpoint reached by deterministic continuation from time $t_i$. Before freezing, different times may lead to distinct continuation endpoints; after freezing, later continuations instead relax to mutually similar endpoints.
We quantify this endpoint similarity using the Jaccard similarity between their misclassified test sets,
\begin{equation}
    \mathrm{Sim}_J(\hat{\bm{\theta}}_i,\hat{\bm{\theta}}_j)
    =
    \frac{|\mathcal{E}(\hat{\bm{\theta}}_i)\cap\mathcal{E}(\hat{\bm{\theta}}_j)|}
    {|\mathcal{E}(\hat{\bm{\theta}}_i)\cup\mathcal{E}(\hat{\bm{\theta}}_j)|},
    \label{eq:jaccard_similarity}
\end{equation}
where $\mathcal{E}(\hat{\bm{\theta}})$ is the set of test samples misclassified by the continuation endpoint $\hat{\bm{\theta}}$. Unlike similarity measures defined directly in weight space, which suffer from permutation symmetry of neurons, this error-based similarity is invariant to such symmetries and thus provides a more faithful measure of solution similarity.

The resulting similarity matrix [Fig.~\ref{fig:phenomenology}(d)] shows a clear transition: continuation endpoints from early times can differ substantially, whereas endpoints from later times become highly similar, forming the red block in Fig.~\ref{fig:phenomenology}(d). We define the freezing time as the earliest time after which all subsequent continuation endpoints remain mutually similar,
\begin{equation}
t_f
\equiv
\min\left\{
t_c \ \middle|\
\mathrm{Sim}_J(\hat{\bm{\theta}}_i,\hat{\bm{\theta}}_j)\ge S_{\mathrm{th}},
\;\; \forall\; t_i,t_j \geq t_c
\right\},
\label{eq:freezing_time}
\end{equation}
where $S_{\mathrm{th}}$ is a high-similarity threshold. The definition is robust to the choice of $S_{\mathrm{th}}$ and remains strongly correlated with alternative freezing criteria based on prediction agreement or a training-loss proxy (see Supplemental Material for details). Note that this basin-identity freezing time is distinct from the time at which training accuracy first reaches $100\%$: the latter only marks perfect classification of the training set, whereas $t_f$ marks stable commitment to the final basin; in most hyperparameter conditions, this basin commitment occurs no later than perfect training accuracy (Fig.~S4 in the Supplemental Material).

Before the freezing time $t_f$, the continuation endpoints become progressively flatter and achieve lower test loss. This trend is reproduced across independent repeats: averaging over 20 SGD trajectories at the same hyperparameter setting ($B=50$, $\eta=0.05$), the endpoint test loss decreases while the endpoint flatness increases as the continuation time $t_c$ approaches and passes the mean freezing time $\langle t_f\rangle$ [Fig.~\ref{fig:phenomenology}(e)]. After this transition, both quantities vary more weakly with $t_c$, indicating that later continuations relax to similar high-flatness endpoints. Thus, the key transition is not final convergence itself, but the earlier point at which SGD loses the ability to move between competing valleys. Together, these results show that transient valley hopping drives SGD toward flatter minima before the trajectory becomes frozen.

\subsection{SGD noise delays the freezing transition}
\begin{figure}[!tbp]
\centering
\includegraphics[width=\linewidth]{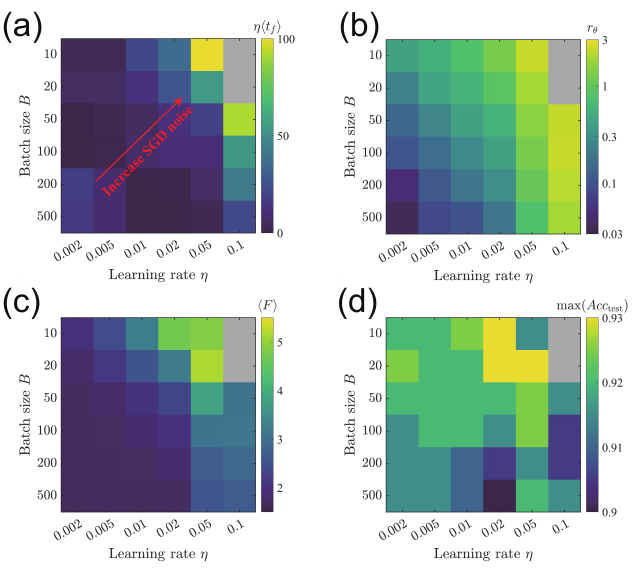}
\caption{SGD noise delays freezing and broadens access to flatter final solutions. (a) Normalized mean freezing time $\eta \langle t_f \rangle$ across learning rates and batch sizes. 
(b) Across-repeat root-mean-square endpoint radius $r_\theta$ for repeats that reached final convergence, shown on a logarithmic color scale. 
(c) Mean final flatness $\langle F \rangle$ for the same converged repeats. 
(d) Best-of-repeats final test accuracy, $\max({Acc}_{\mathrm{test}})$. 
Gray cells indicate settings with no final-converged repeats.}
\label{fig:freezing_control}
\end{figure}

Having identified the freezing time along a single trajectory, we next examined how it depends on the strength of SGD noise. In mini-batch SGD, stochasticity comes from finite-batch sampling, with an effective strength controlled by the learning rate $\eta$ and batch size $B$: larger $\eta$ or smaller $B$ gives stronger noise, with an effective scale $\eta/B$ \cite{jastrzebskiThreeFactorsInfluencing2018}. To test this dependence, we trained networks over a broad grid of $(\eta,B)$, with learning rates from $\eta=0.002$ to $0.1$ and batch sizes from $B=10$ to $500$. For each setting, we performed 20 independent runs starting from the same initialization and measured the freezing time using matched continuation (see Supplemental Material for details). Across stochastic settings with $B<1000$, the normalized freezing time $\eta \langle t_f \rangle$ increases with the effective noise scale $\eta/B$ [Fig.~\ref{fig:freezing_control}(a)], where $\eta$ rescales iteration number to optimization time. Thus, stronger SGD noise delays freezing and extends the transient period during which the trajectory can still move between competing valleys.

This delayed freezing also affects the set of final solutions reached across repeats. For each hyperparameter setting, we measured the root-mean-square (RMS) radius of the converged endpoints,
\begin{equation} r_\theta \equiv \left[ \left\langle \left\| \hat{\bm{\theta}}^{(n)} - \langle\hat{\bm{\theta}}^{(n)}\rangle_n \right\|^2 \right\rangle_n \right]^{1/2}, 
\end{equation}
where $\hat{\bm{\theta}}^{(n)}$ denotes the final endpoint of repeat $n$. Stronger-noise settings produce larger $r_\theta$ [Fig.~\ref{fig:freezing_control}(b)], showing that delayed freezing is associated with a broader set of accessible endpoints. The same ordering is observed for the mean final flatness [Fig.~\ref{fig:freezing_control}(c)], suggesting that this broader set is biased toward flatter regions. Although test accuracy varies only weakly in this small task, the accessibility towards broader and flatter endpoints increases the chance of finding a more generalizable solution, as showin by the improved best-of-repeats test accuracy in Fig.~\ref{fig:freezing_control}(d).

Together, these results clarify the dual role of the SGD noise. Stronger SGD noise not only drives valley-hopping it also delays the freezing transition and thus extends the exploration phase favoring flatter minima.

\subsection{A minimal model of valley hopping and freezing}

\begin{figure}[tbhp]
\centering
\includegraphics[width=\linewidth]{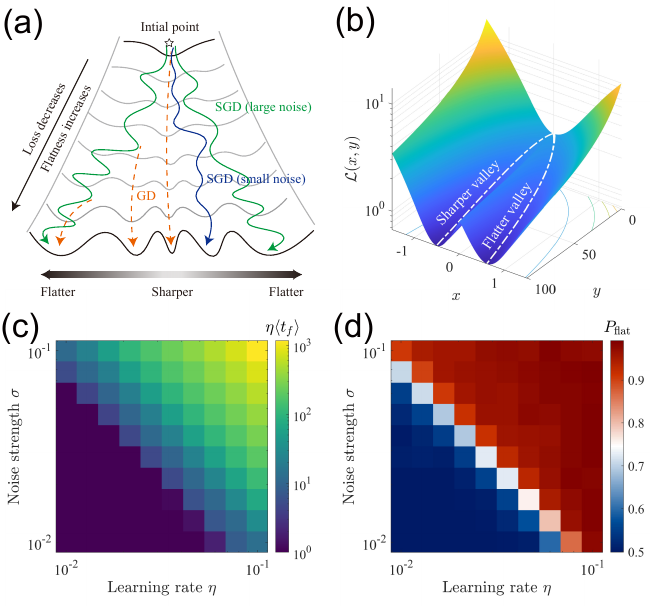}
\caption{A minimal two-valley model with anisotropic, landscape-dependent noise recapitulates the empirically observed training dynamics. 
(a) Conceptual schematic of key characteristics of empirical SGD dynamics. 
(b) The constructed 2D loss landscape $\mathcal{L}(x, y)$ with bifurcating sharper and flatter valleys of equal depth, as defined in Eq.~\ref{eq:loss_toy_model}. 
(c and d) Heatmaps showing the normalized mean freezing time $\eta \langle t_f \rangle$ (c) and the probability of converging to the flatter valley $P_{\mathrm{flat}}$ (d), as functions of learning rate $\eta$ and noise strength $\sigma$.
Statistics were collected from 2000 simulations, half initialized on the sharper side and half on the flatter side for each hyperparameter setting. Full simulation procedures and landscape parameters are provided in the Supplemental Material, Sec.~II~A.
}
\label{fig:toy_model}
\end{figure}

The above results suggest a dynamical picture of how SGD selects among competing solutions [Fig.~\ref{fig:toy_model}(a)]. As training proceeds, the trajectory encounters multiple solution valleys with different flatness, and the accessible valleys become increasingly flatter along the training direction. During an early transient phase, SGD can hop between these valleys before freezing to one of them. Crucially, stronger SGD noise delays this freezing, thereby extending the exploratory window and broadening access to flatter valleys.

To capture these phenomena analytically, we construct a two-dimensional loss landscape $\mathcal{L}(x, y)$ with a bifurcating valley [Fig.~\ref{fig:toy_model}(b)]. Here, $y$ represents the main training direction along which the overall loss decreases and valley flatness increases, while $x$ denotes a symmetry-breaking direction along which the valley splits. The loss is defined piecewise as
\begin{equation}
    \mathcal{L}(x, y) = \mathcal{L}_0(y)  +
    \begin{cases}
       \dfrac{x(x - g_1(y))}{f_1(y)}, & x \geq 0, \\
       \dfrac{x(x + g_2(y))}{f_2(y)}, & x < 0,
    \end{cases}
    \label{eq:loss_toy_model}
\end{equation}
where $g_{1,2}(y)$ set the valley-minimum positions that bifurcate from the origin and asymptotically approach $\pm x_{1,2}$, $f_{1,2}(y)$ control the valley curvatures (flatness), and $\mathcal{L}_0(y)$ provides a global drift capturing the overall loss reduction during training. The landscape is constructed so that both valleys have equal depth relative to the barrier at $x=0$, isolating the geometric effect of flatness from valley depth. The flatness ratio $\gamma \equiv f_1/f_2 = (x_1/x_2)^2 > 1$ ensures that the $x>0$ valley is always flatter. As $y$ increases, both valleys become flatter while the inter-valley barrier height $\Delta\mathcal{L}(y)$ grows, providing the conditions for transient freezing. Full definitions of the landscape functions and parameter values are given in Appendix~\ref{sec:appendix0} and Supplemental Material, Sec.~II~A.

The learning dynamics follow the discrete-time update rule
\begin{equation}
    \bm{\theta}_{t+1} = \bm{\theta}_t - \eta \bigl( \nabla \mathcal{L}(\bm{\theta}_t) + \bm{\xi}_t \bigr),
\label{eq:SGD_updating_rule}
\end{equation}
where $\bm{\xi}_t \sim \mathcal{N}(0, 2\sigma \mathbf{H}(\bm{\theta}_t))$ is Hessian-aligned anisotropic noise with strength $\sigma$ inversely dependent on batch size~\cite{fengInverseVarianceFlatness2021,zhuAnisotropicNoiseStochastic2019,xieDiffusionTheoryDeep2021,liHessianBasedAnalysis2020,yangStochasticGradientDescent2023}. The qualitative behavior is insensitive to the exact functional form; any monotonically increasing relationship between $\mathbf{\Sigma}$ and $\mathbf{H}$ yields the same qualitative results (see Sec.~II~B in Supplemental Material and Fig.~S6).

Simulations of this simple model reproduce our key empirical findings. Starting from the two valleys with equal probabilities, we varied $\eta$ and $\sigma$ and measured the normalized mean freezing time $\eta \langle t_f \rangle$ ($t_f$ defined as the last valley-hopping iteration) and the probability of converging to the flatter valley $P_{\mathrm{flat}}$ (see Sec.~II~A in Supplemental Material). The normalized freezing time increases systematically with both $\eta$ and $\sigma$ [Fig.~\ref{fig:toy_model}(c)], paralleling our neural network experiments [compare with Fig.~\ref{fig:freezing_control}(a)]. Delayed freezing further increases the chance of ending in the flatter valley: as $\eta$ or $\sigma$ increases, $P_{\mathrm{flat}}$ rises sharply [Fig.~\ref{fig:toy_model}(d)], consistent with the empirical finding that stronger SGD noise leads to flatter final solutions [compare with Fig.~\ref{fig:freezing_control}(c)].

Despite its simplicity, the two-valley model captures the essential features of the learning dynamics: the anisotropic landscape-dependent noise not only enables escape from sharper minima but also biases convergence toward flatter regions by extending the early transient exploration phase, thereby allowing the dynamics to settle preferentially into flatter solution valleys. We next develop an analytical theory for valley selection in this model. The analysis proceeds in two steps: we first derive a quasi-steady-state bias toward flatter valleys at fixed $y$, and then show that transient freezing dynamics determine where along the $y$-direction this bias is ultimately frozen in.

\subsection{Anisotropic SGD noise reshapes the effective loss landscape}

\begin{figure*}[t]
\centering
\includegraphics[width=0.98\textwidth]{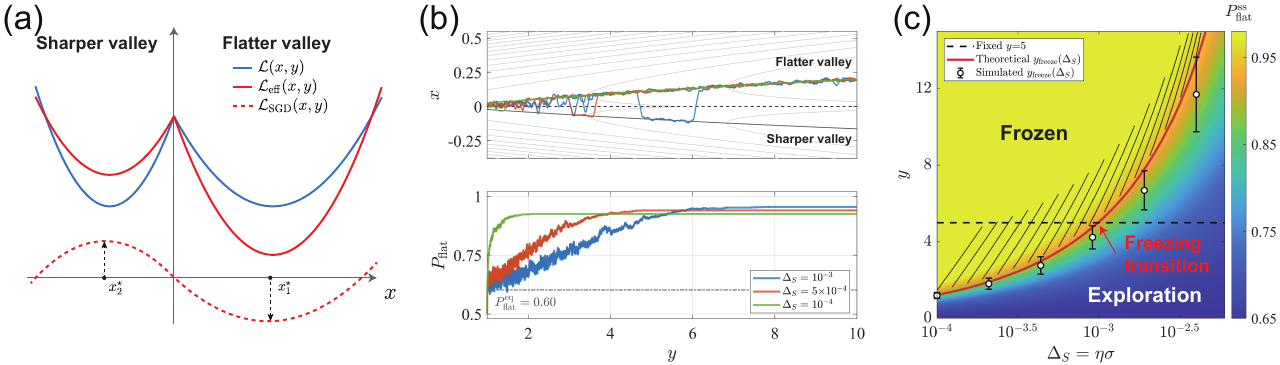}
\caption{Analytical two-valley model for SGD-induced valley selection.
(a) Anisotropic SGD noise reshapes the original loss $\mathcal{L}(x,y)$ (blue solid line) into an effective loss $\mathcal{L}_{\mathrm{eff}}(x,y)$ (red solid line) by adding a correction term $\mathcal{L}_{\mathrm{SGD}}(x,y)$ (red dashed line). 
(b) Representative stochastic trajectories in the two-valley landscape (top) and the corresponding ensemble flat-valley probability $P_{\mathrm{flat}}(y)$ (bottom) for different effective noise strengths $\Delta_S$. In the trajectory plot, the dashed line marks the ridge at $x=0$ and the gray curves show contours of the loss landscape. The equilibrium reference $P_{\mathrm{flat}}^{\mathrm{eq}}= (1+\gamma^{-1})^{-1}$ is marked by the gray dash-dotted line.
(c) Steady-state flat-valley probability $P_{\mathrm{flat}}^{\mathrm{ss,SGD}}(y,\Delta_S)$ from Eq.~\ref{eq:ss_SGD}. The black dashed line marks a fixed-$y$ slice at $y=5$. The red curve indicates the theoretical freezing transition boundary $y_{\mathrm{freeze}}(\Delta_S)$ estimated from Eq.~\ref{eq:y_freeze}, which separates the exploration phase (below) from the frozen phase (above); white circles with error bars show the mean and standard deviation of $y_{\mathrm{freeze}}(\Delta_S)$ from direct simulations.}
\label{fig:analytical_results}
\end{figure*}

In the continuous-time limit (small $\eta$), the SGD dynamics are governed by a Fokker--Planck equation with an anisotropic, landscape-dependent diffusion tensor $\mathbf{D}(x,y) = \Delta_S \mathbf{H}(x,y)$, where $\Delta_S \equiv \eta\sigma$ is the effective noise level (see Appendix~\ref{sec:appendix1}). Because the transverse relaxation timescale $\tau_x$ is much shorter than the longitudinal drift timescale $\tau_y$, the fast variable $x$ rapidly equilibrates at each value of the slowly evolving $y$. In this quasi-steady-state regime, solving the Fokker--Planck equation reveals that the learning dynamics effectively explores a noise-induced effective potential (see Appendix~\ref{sec:appendix2} for derivation):
\begin{equation}
\begin{split}
    \mathcal{L}_{\mathrm{eff}}^{\pm}(x,y)
    &\equiv -T_{\mathrm{eff}}(y)\ln P_\mathrm{ss}^{\pm}(x,y)  \\
    &\approx
    \gamma^{\pm\frac{1}{2}}\mathcal{L}(x,y)
    +\left(1-\gamma^{\pm\frac{1}{2}}\right)\mathcal{L}_0(y),
\end{split}
\label{eq:effective_loss_SGD}
\end{equation}
where $T_{\mathrm{eff}}(y) \equiv 2\Delta_S / \sqrt{f_1(y)f_2(y)}$ is the effective temperature, and the superscripts ``$+$'' and ``$-$'' denote quantities associated with the flatter and sharper valleys, respectively. This effective loss decomposes into the original loss plus an SGD-induced correction:
\begin{equation}
\begin{split}
    \mathcal{L}_{\mathrm{SGD}}^{\pm}(x,y)
    &\equiv \mathcal{L}_{\mathrm{eff}}^{\pm}(x,y)-\mathcal{L}(x,y) \\
    &= \left(\gamma^{\pm\frac{1}{2}}-1\right)
    \big[\mathcal{L}(x,y)-\mathcal{L}_0(y)\big].
\end{split}
\end{equation}
Since $\gamma > 1$, this correction lowers the effective loss near the flat minimum while raising it near the sharp one [Fig.~\ref{fig:analytical_results}(a)], revealing how anisotropic SGD noise reshapes the landscape to favor flatter valleys.

Applying Kramers’ rate theory to estimate the inter-valley escape rates (see Appendix~\ref{sec:appendix3} for details), the steady-state probability of residing in the flat valley is:
\begin{equation}
\begin{split}
    P_{\mathrm{flat}}^{\mathrm{ss,SGD}}(y)
    &= \left[1 + \gamma^{-1}\frac{\mathrm{erfi}\left(\sqrt{\frac{ \Delta \mathcal{L}(y)f_2(y)}{2\Delta_S}}\right)}
                     {\mathrm{erfi}\left(\sqrt{\frac{ \Delta \mathcal{L}(y)f_1(y)}{2\Delta_S}}\right)}\right]^{-1}  \\
    &\approx  \left[1 + \gamma^{-\frac{1}{2}}
    \exp\left(\frac{\Delta \mathcal{L}(y)(f_2(y)-f_1(y))}{2\Delta_S}\right)\right]^{-1},
\label{eq:ss_SGD}
\end{split}
\end{equation}
where $\Delta\mathcal{L}(y)$ is the common barrier height (Appendix~\ref{sec:appendix0}) and $\mathrm{erfi}(\cdot)$ denotes the imaginary error function. Because $\mathrm{erfi}(\cdot)$ is monotonically increasing, $P_{\mathrm{flat}}^{\mathrm{ss,SGD}}$ always exceeds its equilibrium counterpart $P_{\mathrm{flat}}^{\mathrm{eq}} = (1+\gamma^{-1})^{-1}$, confirming that anisotropic SGD noise enhances the preference for the flatter valley, consistent with previous work~\cite{xieDiffusionTheoryDeep2021}.

This prediction is illustrated by the simulations in Fig.~\ref{fig:analytical_results}(b). The ensemble probability $P_{\mathrm{flat}}(y)$ stays above the equilibrium reference $P_{\mathrm{flat}}^{\mathrm{eq}}$ throughout, in agreement with Eq.~\ref{eq:ss_SGD}. Individual trajectories cross the ridge at $x=0$ stochastically at small $y$, but become confined to a single valley as $y$ grows, so that $P_{\mathrm{flat}}(y)$ stops evolving, resulting in the freezing phase. The steady-state result of Eq.~\ref{eq:ss_SGD} therefore applies only while the system remains in the active exploration (valley-hopping) phase; the final valley is selected once inter-valley hopping freezes at a particular value of $y$. We next examine how this freezing point, and hence the final solution, is determined.

\subsection{Noise-delayed freezing selects the final solution}
At fixed $y$, Eq.~\ref{eq:ss_SGD} predicts that increasing $\Delta_S$ weakens the preference for the flatter valley---seemingly contradicting both our empirical results [Fig.~\ref{fig:freezing_control}(c)] and simulations [Fig.~\ref{fig:toy_model}(d)]. This apparent paradox arises because the effective loss (Eq.~\ref{eq:effective_loss_SGD}) is independent of $\Delta_S$; at fixed $y$, raising the effective temperature would reduce the probability of selecting the flatter solution. The resolution lies in the transient dynamics of $y$: the value of $y$ at which Eq.~\ref{eq:ss_SGD} ultimately applies depends on $\Delta_S$.

This distinction is illustrated in Fig.~\ref{fig:analytical_results}(c). Along a fixed-$y$ slice (black dashed line), increasing $\Delta_S$ lowers the steady-state flat-valley probability predicted by Eq.~\ref{eq:ss_SGD}. The transient dynamics relevant for final selection instead follows the freezing boundary (red curve): it does not freeze at a common value of $y$ for all noise strengths, and stronger noise shifts the freezing point to larger $y$. Since $y$ increases monotonically as training proceeds, the freezing time $t_f$ measured in our experiments and simulations maps onto a freezing location $y_{\mathrm{freeze}} \equiv y(t_f)$, in terms of which the transient theory is most naturally formulated.

Early in learning, when barriers are low, inter-valley transitions are frequent and the distribution rapidly equilibrates via Eq.~\ref{eq:ss_SGD}. As $y$ increases, the growing barrier height and the increasing valley flatness progressively suppress escape, until the system reaches a freezing point $y_{\mathrm{freeze}}$ where transitions effectively stop [Fig.~\ref{fig:analytical_results}(b,c)]. The final valley occupation is set by the quasi-steady-state distribution frozen at this point:
\begin{equation}
P_{\mathrm{flat}}^{\mathrm{tr,SGD}} \approx
\left[1+\gamma^{-\frac{1}{2}}
\exp\!\left(\frac{x_2^2-x_1^2}{2 \Delta_S}
\left(\frac{y_\mathrm{freeze}}{y_b+y_\mathrm{freeze}}\right)^2\right)\right]^{-1}.
\label{eq:tr_SGD_original}
\end{equation}
Crucially, $y_\mathrm{freeze}$ increases with $\Delta_S$: stronger noise keeps the system mobile for longer, delaying the freezing point. Within the asymptotic regime where the escape-rate prefactor varies slowly compared with the exponential barrier factor, the freezing criterion gives the estimate (see Appendix~\ref{sec:appendix4} for derivation):
\begin{equation}
y_\mathrm{freeze} \approx y_b \cdot \frac{\sqrt{\Delta_S \ln\left( \frac{\Delta_S}{\varepsilon^2 \Phi^2} \right)}}{x_2 - \sqrt{\Delta_S \ln\left( \frac{\Delta_S}{\varepsilon^2 \Phi^2} \right)}},
\label{eq:y_freeze}
\end{equation}
where $\varepsilon$ is a small constant and $\Phi$ is an effective constant absorbing slowly varying prefactors and other model parameters. This expression is meaningful in the regime $0<\Delta_S \ln[\Delta_S/(\varepsilon^2\Phi^2)]<x_2^2$, where the logarithm is positive and the predicted freezing point remains finite. Substituting Eq.~\ref{eq:y_freeze} into Eq.~\ref{eq:tr_SGD_original} yields:
\begin{equation}
P_{\mathrm{flat}}^{\mathrm{tr,SGD}} \approx \left[ 1 + \gamma^{-\frac{1}{2}} \left( \frac{\sqrt{\Delta_S}}{\varepsilon \Phi} \right)^{1-\gamma} \right]^{-1},
\label{eq:tr_SGD}
\end{equation}
which, since $\gamma > 1$, shows that $P_{\mathrm{flat}}$ increases monotonically with noise strength within the same freezing-approximation regime. In Fig.~\ref{fig:analytical_results}(c), this corresponds to moving along the red freezing transition boundary rather than along a fixed-$y$ slice. Thus, while stronger noise reduces the static preference at fixed $y$, this effect is dominated by delayed freezing: the system explores longer and ultimately settles in the flatter valley with greater probability, resolving the apparent paradox.

\section{Discussion}
We have shown that flat-minimum selection in SGD can be understood as a nonequilibrium freezing problem. Anisotropic, landscape-dependent noise reshapes the loss landscape into an effective potential that biases dynamics toward flatter valleys. Crucially, the final outcome is not set by this static bias alone: the trajectory remains mobile among competing valleys during an early transient phase, and the selected basin is determined at a finite freezing time when inter-valley transitions cease. Stronger noise delays this freezing, extending the exploratory window and increasing the probability of settling in a flatter valley.

This mechanism is fundamentally distinct from previously proposed explanations based on dynamic instability, in which sharp minima become unstable and inaccessible at high learning rates \cite{wuHowSGDSelects2018}. By contrast, our results demonstrate that a robust bias toward flat minima emerges even when all valleys remain dynamically stable. In this regime, anisotropic stochasticity alone is sufficient to drive preferential selection through transient freezing. Our framework therefore explains how SGD favors flat solutions across a broad range of stable training conditions, without invoking instability or divergence.

The role of noise in our framework bears a conceptual resemblance to the Mpemba effect, where a hotter system can freeze faster than a cooler one due to nonequilibrium relaxation pathways~\cite{mpembaCool1969,auerbachSupercoolingMpembaEffect1995,brownridgeWhenDoesHot2011,luNonequilibriumThermodynamicsMarkovian2017}. In both cases, the outcome defies equilibrium intuition and is governed by transient dynamics rather than static properties. However, the mechanism here operates differently: stronger noise delays rather than accelerates freezing, and it is this delayed commitment that steers the optimizer toward a superior frozen state. The relevant question is therefore not how fast the system freezes, but where it freezes---a distinction that highlights the selective, rather than kinetic, role of noise in SGD.

Although our analytical treatment uses a minimal two-dimensional model, the underlying mechanism extends naturally to high-dimensional networks, where optimization is effectively constrained to a low-dimensional submanifold of parameter space. The framework also generalizes to landscapes with many competing valleys: because the bias strengthens with curvature contrast [Fig.~\ref{fig:analytical_results}(c)], transitions are directionally biased toward progressively flatter regions until halted by freezing. While our primary experiments use a multilayer perceptron on MNIST, both multi-valley structure and transient learning dynamics are widely observed across architectures and datasets \cite{fortLargeScaleStructure2019,garipovLossSurfacesMode2018,achilleCriticalLearningPeriods2019,jastrzebskiBreakEvenPointOptimization2020,kalraPhaseDiagramEarly2023}; we provide additional results for a convolutional network on CIFAR-10 in Supplemental Material, Sec.~I~F.

It is worth noting that delayed freezing in normalized optimization time $\eta t_f$ does not imply slower training in practice. Reducing the batch size $B$ increases the number of parameter updates per epoch, while increasing the learning rate $\eta$ enlarges each update step; both effects accelerate per-epoch progress. Consequently, stronger-noise settings can extend the exploratory phase in normalized time---allowing the trajectory to discover flatter valleys---while maintaining or even reducing the total training cost.

More broadly, our results suggest that learning-rate and batch-size schedules should be understood as tools for controlling the timing of basin commitment, not merely for ensuring convergence. This perspective---viewing SGD as a noise-regulated selection process that exploits landscape geometry to find generalizable solutions---opens a path toward principled, dynamics-based design of optimization algorithms grounded in nonequilibrium statistical physics.

\section*{Data and Code Availability}
The code used for the empirical experiments and the two-valley simulations is available at
\url{https://github.com/YangNing1995/Transient_Dynamics_SGD}.

\begin{acknowledgments}
The work by N.Y. and C.T. was supported by the National Natural Science Foundation of China (Grant No. 12505089) and Fundamental and Interdisciplinary Disciplines Breakthrough Plan of the Ministry of Education of China (JYB2025XDXM502). The Flatiron Institute is a division of the Simons Foundation.
\end{acknowledgments}

\appendix

\setcounter{equation}{0}
\renewcommand{\theequation}{A\arabic{equation}}

\section{Analytical results of the two-valley model}

\subsection{Construction of the two-valley loss landscape} \label{sec:appendix0}

The two-valley loss landscape $\mathcal{L}(x,y)$ is given by Eq.~\ref{eq:loss_toy_model} in the main text. The valley-position functions $g_{1,2}(y)$ are:
\begin{equation}
    g_1(y) = \frac{2 x_1 y}{y + y_b}, \qquad
    g_2(y) = \frac{2 x_2 y}{y + y_b},
\end{equation}
so that the valley minima are located at $x_1^{\star} = g_1(y)/2$ and $x_2^{\star} = -g_2(y)/2$, asymptotically approaching $x_1$ and $-x_2$ as $y \to \infty$. The parameter $y_b$ sets the bifurcation scale. The valley flatness functions are:
\begin{equation}
    f_1(y) = f_0 \frac{x_1^2}{x_0^2} \frac{(y+y_f)^2}{(y+y_b)^2},
    \qquad
    f_2(y) = f_0 \frac{x_2^2}{x_0^2} \frac{(y+y_f)^2}{(y+y_b)^2},
\end{equation}
where $f_0$ sets the overall flatness scale, $x_0$ is a normalization constant, and $y_f$ controls the rate at which both valleys flatten as $y$ increases. Their ratio $\gamma \equiv f_1/f_2 = (x_1/x_2)^2$ remains constant. The two valley minima have equal depths relative to the barrier at $x=0$, with barrier height:
\begin{equation}
\begin{split}
    \Delta \mathcal{L}(y)
    &= \mathcal{L}(0, y) - \mathcal{L}(x_1^{\star}, y)
     = \mathcal{L}(0, y) - \mathcal{L}(x_2^{\star}, y)  \\
    &= \frac{x_0^2}{f_0} \frac{y^2}{(y+y_f)^2},
\end{split}
\end{equation}
which increases with $y$ and saturates at $x_0^2/f_0$. Finally, the global drift term is:
\begin{equation}
    \mathcal{L}_0(y) = \mathcal{L}_d e^{-y / y_d} + \mathcal{L}_0^{\star},
\end{equation}
where $\mathcal{L}_d$ controls the initial drift strength, $y_d$ sets its decay scale, and $\mathcal{L}_0^{\star} = x_0^2 / f_0$ ensures $\mathcal{L} \ge 0$ everywhere. Parameter values are given in Table~S1 of the Supplemental Material.

\subsection{Continuous-time dynamics and Fokker-Planck equation} \label{sec:appendix1}

We analyze the continuous-time limit of the discrete SGD update rule:
\begin{equation}
    \bm{\theta}_{t+1} = \bm{\theta}_t - \eta \nabla \mathcal{L}(\bm{\theta}_t) - \eta \bm{\xi}_t,
    \label{eq:si_sgd_update} 
\end{equation}
where the gradient noise satisfies $\mathbb{E}[\bm{\xi}_t] = 0$ and $\text{Cov}[\bm{\xi}_t] = \mathbf{\Sigma}(\bm{\theta}_t)$. By interpreting the learning rate $\eta$ as the discretization time step $\Delta t$, the process converges to the following stochastic differential equation (SDE) in the limit $\eta \to 0$ \cite{liHessianBasedAnalysis2020,liStochasticModifiedEquations2017}:
\begin{equation}
    \dd\bm{\theta}_t = \mathbf{A}(\bm{\theta}_t) \dd t + \sqrt{2\mathbf{D}(\bm{\theta}_t)} \dd \mathbf{W}_t,
    \label{eq:si_sde} 
\end{equation}
where $\dd \mathbf{W}_t$ denotes the standard Wiener increment. To determine the drift vector $\mathbf{A}$ and the diffusion tensor $\mathbf{D}$, we match the first and second moments of the discrete update with those of the SDE:

\begin{enumerate}
    \item \textbf{Drift:} Matching the deterministic components, $\mathbf{A}(\bm{\theta}) \Delta t = -\eta \nabla \mathcal{L}(\bm{\theta})$, yields:
    \begin{equation}
        \mathbf{A}(\bm{\theta}) = -\nabla \mathcal{L}(\bm{\theta}).
    \end{equation}
    \item \textbf{Diffusion:} Matching the covariance of the noise terms, $2\mathbf{D}\Delta t = \text{Cov}[-\eta \bm{\xi}]$, implies $2\mathbf{D}\eta = \eta^2 \mathbf{\Sigma}$. This gives:
    \begin{equation}
        \mathbf{D}(\bm{\theta}) = \frac{1}{2} \eta \mathbf{\Sigma}(\bm{\theta}).
    \end{equation}
\end{enumerate}

Under the assumption that the noise covariance is proportional to the Hessian, $\mathbf{\Sigma}(\bm{\theta}) = 2\sigma \mathbf{H}(\bm{\theta})$, the diffusion tensor takes the form:
\begin{equation}
    \mathbf{D}(\bm{\theta}) = \eta \sigma \mathbf{H}(\bm{\theta}) \equiv \Delta_S \mathbf{H}(\bm{\theta}),
    \label{eq:si_D_final}
\end{equation}
where $\Delta_S \equiv \eta\sigma$ represents the effective noise intensity.

The time evolution of the probability density $P(\bm{\theta}, t)$ is governed by the continuity equation $\partial_t P = -\nabla \cdot \mathbf{J}$ \cite{gardiner2009stochastic}. Substituting the drift $\mathbf{A} = -\nabla \mathcal{L}$ and diffusion $\mathbf{D}$ into the probability current $\mathbf{J} = \mathbf{A}P - \mathbf{D}\nabla P$, we obtain the Fokker-Planck equation:
\begin{equation}
    \partial_t P(\bm{\theta}, t) = \nabla \cdot \Big[ P(\bm{\theta}, t) \nabla \mathcal{L}(\bm{\theta}) 
    + \mathbf{D}(\bm{\theta}) \cdot \nabla P(\bm{\theta}, t) \Big].
    \label{eq:si_fp_final} 
\end{equation}

\subsection{Non-equilibrium steady state and effective loss} \label{sec:appendix2}

Our derivation relies on the adiabatic approximation, the validity of which depends on a distinct separation of timescales between the fast transverse relaxation in $x$-direction and the slow longitudinal drift $y$-direction.

The dynamics in the $x$-direction are governed by the local curvature $\kappa_x \equiv \partial_x^2 \mathcal{L}$. The characteristic transverse relaxation time $\tau_x$, representing the time required for the system to return to local equilibrium, is estimated as the inverse of this curvature:
\begin{equation}
    \tau_x(y) \approx \left( \frac{\partial^2 \mathcal{L}}{\partial x^2} \right)^{-1} = \frac{1}{2} f_{1(2)}(y).
\end{equation}
Using the parameters $f_0=x_0=1.0$ and $x_1=0.8$ from the simulation (Table.~S1 of SM), the relaxation time is bounded by $\tau_x^{\mathrm{max}} \approx 0.32$, which indicates a timescale of order $\mathcal{O}(10^{-1})$. Conversely, the evolution of the slow variable $y$ is primarily driven by the global drift. The longitudinal drift timescale $\tau_y$ is given by:
\begin{equation}
    \tau_y(y) = \left| \frac{\dd y}{\dd t} \right|^{-1}\approx \left[ \frac{\mathcal{L}_d}{y_d} e^{-y/y_d} +\frac{2x_0^2 y_f y}{f_0(y+y_f)^3} \right]^{-1}.
\end{equation}
With the simulation parameters, the minimum drift timescale is $\tau_y^{\mathrm{min}} \approx 41.8$. Comparing these conservative estimates yields a ratio of $\tau_x / \tau_y \sim 7.6 \times 10^{-3} \ll 1$. This vast separation justifies the adiabatic approximation.

Furthermore, given the block structure of the Hessian $\mathbf{H}$, the diagonal elements dominate. Specifically, the cross-terms $H_{12}$ are of order $\mathcal{O}(1/y_{f,b})$ and $H_{22}$ is of order $\mathcal{O}(1/y_{f,b}^2)$, whereas $H_{11}$ is of order $\mathcal{O}(1)$. Consequently, we approximate the diffusion matrix as diagonal, keeping only the dominant term in the $x$-direction:
\begin{equation}
    D_{11}(y) \approx \Delta_S H_{11}(x,y) = 
    \begin{cases}
        {2\Delta_S}/{f_1(y)}, & x \geq 0, \\
        {2\Delta_S}/{f_2(y)}, & x < 0.
    \end{cases}
\end{equation}
Neglecting cross-diffusion terms, the Fokker-Planck equation for the conditional distribution in $x$ simplifies to:
\begin{equation}
    \partial_t P(x|y) \approx \partial_x \left[ P(x|y) \partial_x \mathcal{L}(x,y) + D_{11}(y) \partial_x P(x|y) \right].
\end{equation}
Setting the probability flux to zero, the steady-state solution in each valley follows the Boltzmann form:
\begin{equation}
    P_{\mathrm{ss}}^{\pm}(x|y) = \frac{1}{Z_{\mathrm{ss}}^{\pm}(y)} \exp\left( -\frac{\mathcal{L}(x,y)}{D_{11}^{\pm}(y)} \right),
\end{equation}
where the superscripts $+$ and $-$ denote the domains $x \ge 0$ (flatter valley) and $x < 0$ (sharper valley), respectively. Note that the effective temperature $D_{11}^{\pm}$ differs between the two valleys due to the Hessian-dependent noise.

Continuity of the probability density at the junction $x=0$ requires $P_{\mathrm{ss}}^+(0|y) = P_{\mathrm{ss}}^-(0|y)$. The validity of this condition requires inter-valley hopping to occur on a much faster timescale than $\tau_y$, which is maintained before the onset of freezing. Therefore, with $\mathcal{L}(0, y) = \mathcal{L}_0(y)$, this condition implies the following relationship between the normalization constants:
\begin{equation} \label{eq:norm_relation}
\frac{Z_{\mathrm{ss}}^{+}(y)}{Z_{\mathrm{ss}}^{-}(y)}
= \exp\left[
-\mathcal{L}_{0}(y)\left( \frac{1}{D_{11}^{+}(y)} - \frac{1}{D_{11}^{-}(y)} \right)
\right].
\end{equation}
Assuming the drift in the $y$-direction is sufficiently slow, the system reaches a non-equilibrium steady state in both directions. The marginal steady-state probability $P_{\mathrm{ss}}(y)$ is determined by the lowest-order approximation:
\begin{equation}
\begin{split}
    \frac{\dd}{\dd y}\bigg[&\langle \partial_y \mathcal{L} \rangle_x+\frac{\dd}{\dd y} \langle D_{22} \rangle_x \\
    &+\langle D_{22}\rangle_x \frac{\dd}{\dd y} \ln P_{\mathrm{ss}}(y)\bigg] P_{\mathrm{ss}}(y)=0,
\end{split}
\end{equation}
where $\langle \cdot \rangle_x$ denotes integration over the conditional steady-state distribution $P_{\mathrm{ss}}^{\pm}(x|y)$. The final joint steady-state probability $P_{\mathrm{ss}}(x,y)$ is then given by:
\begin{equation}
\begin{split}
    P_{\mathrm{ss}}^{\pm}(x,y)&\equiv P_{\mathrm{ss}}^{\pm}(x|y)P_{\mathrm{ss}}(y) \\
    &=\frac{1}{Z_{\mathrm{ss}}(y)} \exp \left( - \frac{\mathcal{L}(x,y)-\mathcal{L}_{0}(y)}{D_{11}^{\pm}(y)} \right),
\end{split}
\end{equation}
where we have utilized the normalization relationship Eq.~(\ref{eq:norm_relation}) and absorbed all terms independent of $x$ into $Z_{\mathrm{ss}}(y)$.

To interpret this non-equilibrium steady state within the framework of equilibrium statistical mechanics, we define an effective loss $\mathcal{L}_{\mathrm{eff}}(x,y)$. We introduce a global effective temperature $T_{\mathrm{eff}}(y)$ for a fixed $y$, defined as the geometric mean of the diffusion coefficients in the two valleys:
\begin{equation}
    T_{\mathrm{eff}}(y) \equiv \sqrt{D_{11}^+(y) D_{11}^-(y)} = \frac{2\Delta_S}{\sqrt{f_1(y) f_2(y)}}.
\end{equation}
The effective loss is defined via the Boltzmann relation $P_{\mathrm{ss}}(x,y) \propto \exp(-\mathcal{L}_{\mathrm{eff}}(x,y) / T_{\mathrm{eff}})$. Substituting the derived expression for $P_{\mathrm{ss}}(x,y)$, we obtain:
\begin{equation}
\begin{split}
    \mathcal{L}_{\mathrm{eff}}^{\pm}(x,y) &\equiv - T_{\mathrm{eff}}\ln P^{\pm}_{\mathrm{ss}}(x,y) \\
    &= \frac{T_{\mathrm{eff}}}{D_{11}^{\pm}}\left[\mathcal{L}(x,y)-\mathcal{L}_{0}(y)\right] \\
    &\quad +\mathcal{L}_{0}(y)+T_{\mathrm{eff}}\ln Z_{\mathrm{ss}}(y).
\end{split}
\end{equation}
By imposing the condition $\mathcal{L}_{\mathrm{eff}}(0,y) = \mathcal{L}_{0}(y)$, regardless the uniform term $T_{\mathrm{eff}}\ln Z_{\mathrm{ss}}(y)$ for fixed $y$, we recover Eq.~\ref{eq:effective_loss_SGD} of the main text:
\begin{equation}
\begin{split}
    \mathcal{L}_{\mathrm{eff}}^{\pm}(x,y)
    &\approx 
    \sqrt{\frac{f_{1(2)}}{f_{2(1)}}}\,\mathcal{L}(x,y)
    +\left(1-\sqrt{\frac{f_{1(2)}}{f_{2(1)}}}\right)\mathcal{L}_0(y)\\
    &=\gamma^{\pm\frac{1}{2}}\mathcal{L}(x,y)
    +\left(1-\gamma^{\pm\frac{1}{2}}\right)\mathcal{L}_0(y).
\end{split}
\end{equation}
We can thus decompose the effective loss into the original loss plus an SGD-induced correction term:
\begin{equation}
\begin{split}
    \mathcal{L}_{\mathrm{SGD}}^{\pm}(x,y)
    &\equiv \mathcal{L}_{\mathrm{eff}}^{\pm}(x,y)-\mathcal{L}(x,y) \\
    &= \left(\gamma^{\pm\frac{1}{2}}-1\right)
    \big[\mathcal{L}(x,y)-\mathcal{L}_0(y)\big].
\end{split}
\end{equation}
Note that inside the valleys, the term $[\mathcal{L}(x,y) - \mathcal{L}_0(y)]$ is strictly negative. For the flatter valley ($x > 0$), the condition $\gamma\equiv f_1/f_2>1$ implies a positive prefactor $(\gamma^{1/2} - 1) > 0$. Consequently, the SGD correction $\mathcal{L}_{\mathrm{SGD}}^+ < 0$, which effectively deepens the potential well. Conversely, for the sharper valley ($x < 0$), the prefactor becomes negative since $(\gamma^{-1/2} - 1) < 0$. This results in a positive correction $\mathcal{L}_{\mathrm{SGD}}^- > 0$, which effectively raises the potential well.

\subsection{Kramers' escape rates and steady-state convergence probability} \label{sec:appendix3}

We quantify inter-valley transitions using the Mean First Passage Time (MFPT) \cite{gardiner2009stochastic}. Based on the Fokker-Planck equation, the MFPT from the sharp valley minimum ($x_2^{\star}$) to the barrier ($x=0$), denoted $\tau_{s \to f}$, is:
\begin{equation} \label{eq:MFPT_left}
    \tau_{s \to f} = \frac{1}{D_{11}^-} \int_{x_2^{\star}}^{0} \dd x^{\prime} e^{\frac{\mathcal{L}(x^{\prime})}{D_{11}^-}} \int_{-\infty}^{x^{\prime}} \dd x^{\prime\prime} e^{-\frac{\mathcal{L}(x^{\prime\prime})}{D_{11}^-}}.
\end{equation}
Similarly, for the transition from the flat valley minimum ($x_1^{\star}$) to the sharp side ($\tau_{f \to s}$):
\begin{equation} \label{eq:MFPT_right}
    \tau_{f \to s} = \frac{1}{D_{11}^+} \int_{0}^{x_1^{\star}} \dd x^{\prime} e^{\frac{\mathcal{L}(x^{\prime})}{D_{11}^+}} \int_{x^{\prime}}^{+\infty} \dd x^{\prime\prime} e^{-\frac{\mathcal{L}(x^{\prime\prime})}{D_{11}^+}}.
\end{equation}

We apply Kramers' approximation, assuming rare escape events. This regime holds when the noise amplitude is small relative to the barrier height ($D_{11}^\pm \ll \Delta \mathcal{L}$), satisfying:
\begin{equation}
    \Delta_S \ll \frac{1}{2} \left(\frac{x_{1(2)}y}{y+y_{b}}\right)^2.
\end{equation}
In this limit, the double integrals decouple. The inner integral is evaluated via the Gaussian approximation around the potential minimum ($x^{\prime\prime} \approx x_2^{\star}$):
\begin{equation}
\begin{split}
    \int_{-\infty}^{+\infty} \dd x^{\prime\prime} &\exp\left({-\frac{1}{D_{11}^-}\left[\frac{x^{\prime\prime}(x^{\prime\prime}+g_2)}{f_2} + \mathcal{L}_0 \right]}\right) \\
    &\approx \sqrt{\pi D_{11}^-f_2} e^{-\mathcal{L}(x_2^{\star},y)/D_{11}^-}.
\end{split}
\end{equation}
The outer integral is dominated by the region near the barrier top ($x^{\prime} \to 0$), yielding:
\begin{equation}
\begin{split}
   \int_{x_2^{\star}}^{0} \dd x^{\prime} &\exp\left({\frac{1}{D_{11}^-}\left[\frac{x^{\prime}(x^{\prime}+g_2)}{f_2} + \mathcal{L}_0 \right]}\right) \\
   &= \frac{\sqrt{\pi D_{11}^- f_2}}{2}  e^{\mathcal{L}(x_2^{\star},y)/D_{11}^-} \,\mathrm{erfi}\left(\sqrt{\frac{\Delta\mathcal{L}}{D_{11}^{-}}}\right),
\end{split}
\end{equation}
where $\mathrm{erfi}(z)$ is the imaginary error function. Applying the same derivation to $\tau_{f \to s}$, the analytical MFPTs are:
\begin{equation} \label{eq:MFPT_solutions}
    \begin{aligned} 
    \tau_{f \to s} &\approx  \frac{\pi}{2} f_1(y) \, \mathrm{erfi}\left[\sqrt{\frac{\Delta \mathcal{L}(y)f_1(y)}{2\Delta_S}}\right], \\
    \tau_{s \to f} &\approx \frac{\pi}{2} f_2(y) \, \mathrm{erfi}\left[\sqrt{\frac{\Delta \mathcal{L}(y)f_2(y)}{2\Delta_S}}\right].
    \end{aligned}
\end{equation}
Using the asymptotic expansion $\mathrm{erfi}(z) \approx e^{z^2}/(\sqrt{\pi} z)$ for $z \gg 1$, the escape rates $k_\mathrm{ss}^{\pm} \equiv 1/\tau_{f(s) \to s(f)}$ become:
\begin{equation} \label{eq:escape_rates}
    \begin{aligned} 
    k_\mathrm{ss}^{+} &\approx \sqrt{\frac{2 \Delta \mathcal{L}}{\pi \Delta_S f_1}} \exp\left[- \frac{\Delta \mathcal{L} f_1}{2\Delta_S} \right], \\
    k_\mathrm{ss}^{-} &\approx \sqrt{\frac{2 \Delta \mathcal{L}}{\pi \Delta_S f_2}} \exp\left[- \frac{\Delta \mathcal{L} f_2}{2\Delta_S} \right].
    \end{aligned}
\end{equation}
The probability evolution follows the master equation:
\begin{equation} 
    \begin{aligned}
    \frac{\dd P_\mathrm{flat}}{\dd t} &= k_\mathrm{ss}^{-} P_\mathrm{sharp} - k_\mathrm{ss}^{+} P_\mathrm{flat}, \\
    \frac{\dd P_\mathrm{sharp}}{\dd t} &= k_\mathrm{ss}^{+} P_\mathrm{flat} - k_\mathrm{ss}^{-} P_\mathrm{sharp}.
    \end{aligned}
\end{equation}
The steady-state probability of residing in the flat valley, $P_{\mathrm{flat}}^{\mathrm{ss,SGD}} = k_\mathrm{ss}^{-}/(k_\mathrm{ss}^{-} + k_\mathrm{ss}^{+})$, is given by:
\begin{equation}
\label{eq:ss_SGD_a}
\begin{split}
    P_{\mathrm{flat}}^{\mathrm{ss,SGD}}
  &= \left[1 + \gamma^{-1}\frac{\mathrm{erfi}\left(\sqrt{\frac{ \Delta \mathcal{L}f_2}{2\Delta_S}}\right)}
                      {\mathrm{erfi}\left(\sqrt{\frac{ \Delta \mathcal{L}f_1}{2\Delta_S}}\right)}\right]^{-1} \\
    &\approx  \left[1 + \gamma^{-\frac{1}{2}}
    \exp\left(\frac{\Delta \mathcal{L}(f_2-f_1)}{2\Delta_S}\right)\right]^{-1}.
\end{split}
\end{equation}
For isotropic noise ($D_{11}^+ = D_{11}^-$), the effective temperature is uniform, and Eq.~(\ref{eq:ss_SGD}) reduces to the equilibrium distribution:
\begin{equation}
\label{eq:ss_eq_a}
P_{\mathrm{flat}}^{\mathrm{eq}}
= \left(1 + \frac{f_2(y)}{f_1(y)}\right)^{-1}=\frac{\gamma}{1+\gamma}.
\end{equation}
Comparing Eq.~(\ref{eq:ss_SGD_a}) and Eq.~(\ref{eq:ss_eq_a}) confirms that SGD noise introduces a significant bias ($P_{\mathrm{flat}}^{\mathrm{ss,SGD}} > P_{\mathrm{flat}}^{\mathrm{eq}}$) due to the super-linear growth of $\mathrm{erfi}(z)$, amplifying geometric differences between valleys.

\subsection{The freezing mechanism and transient dynamics approximation} \label{sec:appendix4}

The final solution selection is governed by the competition between inter-valley diffusion and longitudinal drift. As $y$ increases, the escape rates $k^{\pm}$ decrease exponentially. The system ``freezes'' when the transition timescale exceeds the drift timescale. The final probability is approximated by the quasi-steady distribution at the freezing point $y_\mathrm{freeze}$:
\begin{equation}
\label{eq:tr_SGD_original_a}
\begin{split}
    P_{\mathrm{flat}}^{\mathrm{tr,SGD}}
    &\approx \left[1 + \gamma^{-\frac{1}{2}} \exp\left( \frac{\Delta \mathcal{L} [f_2 - f_1]}{2 \Delta_S} \right) \right]^{-1} \bigg|_{y_\mathrm{freeze}} \\
    &= \left[1+\gamma^{-\frac{1}{2}} \exp\!\left(\frac{x_2^2-x_1^2}{2 \Delta_S} \left(\frac{y_\mathrm{freeze}}{y_b+y_\mathrm{freeze}}\right)^2\right)\right]^{-1},
\end{split}
\end{equation}
where we have substituted the definitions of $\Delta \mathcal{L}$, $f_1$ and $f_2$.

We define the freezing point $y_{\mathrm{freeze}}$ as the juncture where the escape rate from the sharper valley becomes significantly smaller than the rate of change of the steady-state distribution ($k_\mathrm{ss}^{-} \approx \varepsilon |\dot P/P|$):
\begin{equation}
 k_\mathrm{ss}^{-} \approx -\varepsilon\left(1+\frac{k_\mathrm{ss}^{+}}{k_\mathrm{ss}^{-}}\right)^{-1}\frac{\mathrm{d}}{\mathrm{d}t}\left(\frac{k_\mathrm{ss}^{+}}{k_\mathrm{ss}^{-}}\right),
    \label{eq:freezing_criterion}
\end{equation}
where $\varepsilon$ is a small constant. We expand the time derivative term using the chain rule:
\begin{equation}
\begin{split}
    \frac{\mathrm{d}}{\mathrm{d}t} \left( \frac{k_\mathrm{ss}^{+}}{k_\mathrm{ss}^{-}} \right)
    &\approx \frac{k_\mathrm{ss}^{+}}{2\Delta_S k_\mathrm{ss}^{-}} \frac{\mathrm{d}}{\mathrm{d}y}\Bigl[ \Delta\mathcal{L} \bigl( f_2-f_1 \bigr) \Bigr] \dot{y} \\
    &\approx -\frac{k_\mathrm{ss}^{+}}{2\Delta_S k_\mathrm{ss}^{-}} \left[ \frac{\mathcal{L}_d}{y_d} e^{-y/y_d} +\dots \right] \\
    &\quad \times \left[ \frac{(x_1^2-x_2^2)y_b y}{(y+y_b)^3} \right].
\end{split}
\end{equation}

Inserting this derivative back into the freezing criterion Eq.~(\ref{eq:freezing_criterion}) yields:
\begin{equation}
     k_\mathrm{ss}^{-} \approx \frac{\varepsilon \Phi}{\Delta_S} \frac{k_\mathrm{ss}^{+}}{k_\mathrm{ss}^{+} + k_\mathrm{ss}^{-}}.
\end{equation}
Here, slowly varying prefactors and model parameters are absorbed into $\Phi$. We focus on the regime where freezing occurs well before 
$P_{\mathrm{flat}}^{\mathrm{tr,SGD}}$ reaches 1, so that 
$k_\mathrm{ss}^{-}/k_\mathrm{ss}^{+}=\mathcal{O}(1)$. 
In this regime, the escape rates decay exponentially with $y$, 
much faster than the polynomial decay of $\Phi$. 
We therefore approximate $\Phi$ by its upper bound and treat it as an effective constant:
\begin{equation}
\begin{split}
    \Phi &\equiv \frac{1}{2}\left[ \frac{\mathcal{L}_d}{y_d} e^{-y/y_d} +\dots \right] \left[ \frac{(x_1^2-x_2^2)y_b y}{(y+y_b)^3} \right] \\
    &\approx \frac{2(x_1^2-x_2^2)}{27 y_b}\left(\frac{\mathcal{L}_d}{y_d}+\frac{8x_0^2}{27 y_f}\right).
\end{split}
\end{equation}
This simplifies the relation to ${k_\mathrm{ss}^{-}} \approx \varepsilon \Phi / \Delta_S$. Substituting the explicit exponential form of $k_\mathrm{ss}^{-}$ (Eq.~\ref{eq:escape_rates}), the freezing criterion becomes:
\begin{equation}
    \sqrt{\frac{2 \Delta \mathcal{L}}{\pi \Delta_S f_2}} \exp\left[- \frac{\Delta \mathcal{L} f_2}{2\Delta_S} \right] \approx \frac{\varepsilon \Phi}{\Delta_S}.
\end{equation}
Assuming the prefactor varies slowly, we solve the implicit equation for the exponent:
\begin{equation}
\begin{split}
    \frac{\Delta \mathcal{L}(y_\mathrm{freeze}) f_2(y_\mathrm{freeze})}{2\Delta_S} 
    &\approx \frac{x_2^2}{2\Delta_S} \left(\frac{y_\mathrm{freeze}}{y_b+y_\mathrm{freeze}}\right)^2 \\
    &\approx \ln\left( \frac{\sqrt{\Delta_S}}{\varepsilon \Phi} \right).
\end{split}
\end{equation}
Solving for $y_\mathrm{freeze}$ gives a scaling estimate for its dependence on noise strength:
\begin{equation}
y_\mathrm{freeze} \approx y_b \cdot \frac{\sqrt{\Delta_S \ln\left( \frac{\Delta_S}{\varepsilon^2 \Phi^2} \right)}}{x_2 - \sqrt{\Delta_S \ln\left( \frac{\Delta_S}{\varepsilon^2 \Phi^2} \right)}}.
\label{eq:y_freeze_explicit}
\end{equation}
This estimate applies when $0<\Delta_S \ln[\Delta_S/(\varepsilon^2\Phi^2)]<x_2^2$; outside this range the asymptotic expression either has no real positive freezing point or predicts freezing beyond the finite-$y$ regime captured by the approximation.
Substituting Eq.~(\ref{eq:y_freeze_explicit}) back into Eq.~(\ref{eq:tr_SGD_original_a}), we obtain the final transient probability:
\begin{equation}
\label{eq:final_ptr_corrected}
\begin{split}
    P_{\mathrm{flat}}^{\mathrm{tr,SGD}} 
    &\approx  \left[1+\gamma^{-\frac{1}{2}} \exp\!\left(\frac{x_2^2-x_1^2}{x_2^2} \ln\left( \frac{\sqrt{\Delta_S}}{\varepsilon \Phi} \right)\right)\right]^{-1} \\
    &\approx \left[ 1 + \gamma^{-\frac{1}{2}} \left( \frac{\sqrt{\Delta_S}}{\varepsilon \Phi} \right)^{1-\gamma} \right]^{-1}.
\end{split}
\end{equation}
Since $\gamma = f_1/f_2 > 1$, the exponent $(1-\gamma)$ is negative. Thus, within the freezing-approximation regime, increasing noise $\Delta_S$ reduces the term in the denominator, thereby increasing the probability $P_{\mathrm{flat}}$ of selecting the flatter valley.

\bibliography{deeplearning}

\end{document}